\pdfoutput=1
\documentclass[11pt]{article}

\usepackage[final]{acl}

\usepackage{times}
\usepackage{latexsym}

\usepackage[T1]{fontenc}
\usepackage[utf8]{inputenc}

\usepackage{microtype}

\usepackage{inconsolata}

\usepackage{graphicx}

\usepackage{booktabs}
\usepackage{enumitem}
\usepackage{kotex}
\usepackage{tcolorbox}
\usepackage{adjustbox}
\newcommand{\PAR}[1]{\noindent {\bf #1.~}} 

\title{Responsible Federated LLMs via Safety Filtering and Constitutional AI}

\author{
Eunchung Noh$^{\spadesuit}$\thanks{Equal contribution.}
\quad
Jeonghun Baek$^{\diamondsuit}$\footnotemark[1] \\
$^{\spadesuit}$Samsung Electronics \quad
$^{\diamondsuit}$The University of Tokyo\\
\texttt{eunchung.noh@samsung.com}
}

\begin{document}
\maketitle
\begin{abstract}
Recent research has increasingly focused on training large language models (LLMs) using federated learning, known as FedLLM. However, responsible AI (RAI), which aims to ensure safe and trustworthy responses, remains underexplored in this context. In FedLLM, client-side training data may contain harmful content, resulting in unsafe LLMs that can generate inappropriate responses. Aggregating such models into a global model and redistributing it to clients risks the widespread deployment of unsafe LLMs. To address this, we incorporate two well-established RAI techniques into FedLLM: safety filtering and constitutional AI. Our experiments show that these methods significantly improve LLM safety, achieving over 20\% improvement on AdvBench.
\end{abstract}

\section{Introduction}
Federated learning (FL)~\cite{fedavg,fedavgm,scaffold,fedprox,fedadam} aims to train models using client data while safeguarding client privacy.
To ensure privacy protection, client data remains on local devices rather than being transmitted to a central server.
Instead, the model is trained locally on the client's device, and the locally trained models are then transmitted to the server for aggregation into a global model.
By exchanging model parameters instead of client data, FL enables collaborative learning across distributed clients while preserving data privacy.

Large language models (LLMs)~\cite{GPT3,openai2023gpt4,anil2023palm,team2023gemini,team2024gemma,touvron2023llama1,touvron2023llama2,vicuna2023,alpaca,llama3.1,abdin2024phi,yang2024qwen2} have rapidly advanced and are now widely used in applications such as question answering and chatbots.
Recent research has explored training LLMs via FL, known as FedLLM~\cite{FedPETuning,FedIT,sun2024improving,kuang2024federatedscope,OpenFedLLM}, which aims to finetune LLMs with client data while maintaining privacy.

\begin{figure}[t]
\centering
    \includegraphics[width=\linewidth]{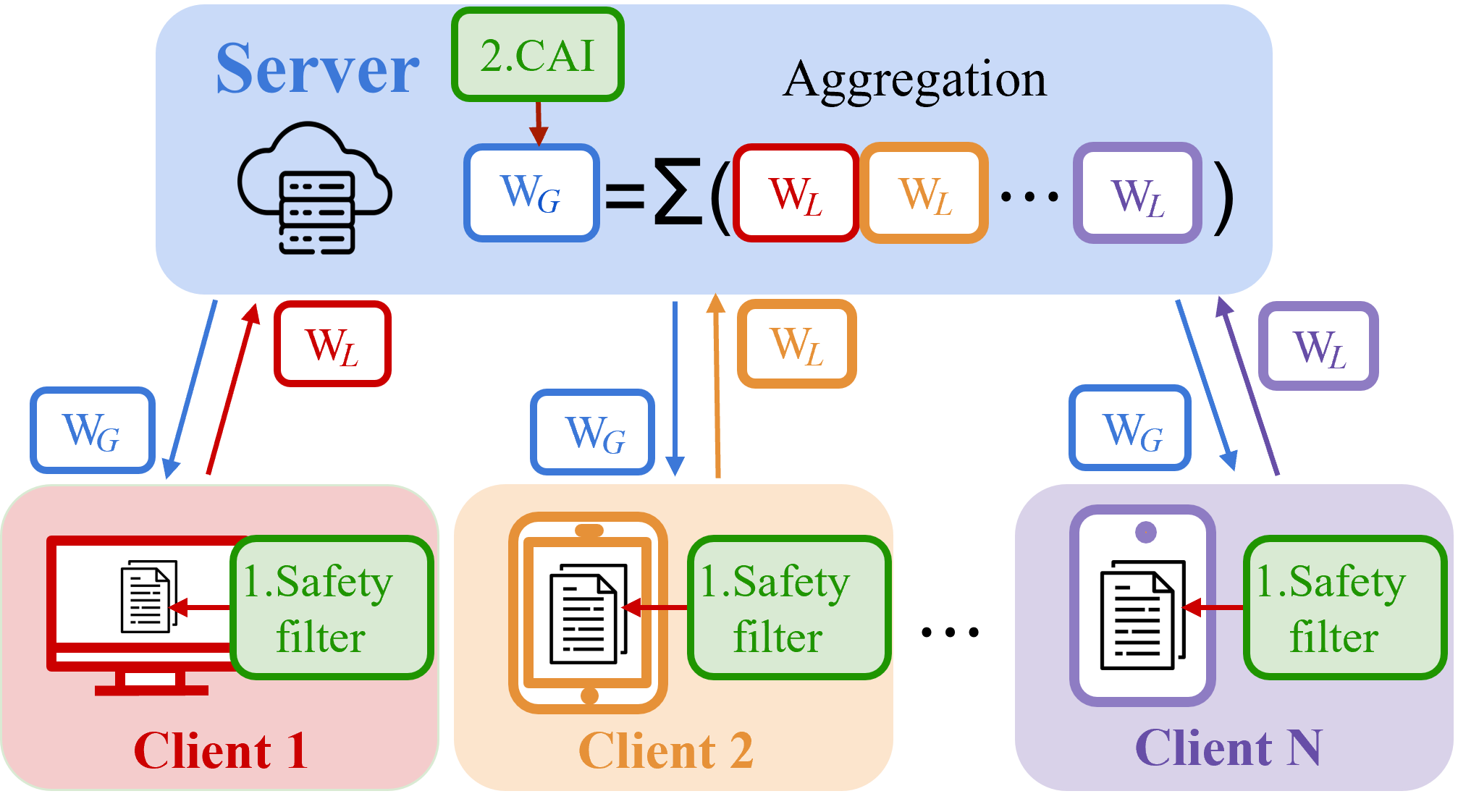} 
    \vspace{-6mm}
    \caption{For the responsible federated large language model (FedLLM), we apply a safety filter and Constitutional AI (CAI) to improve safety.
}
    \label{fig:safetyfilter_CAI}
    % \vspace{-4mm}
\end{figure}

However, previous studies~\cite{FedPETuning,FedIT,sun2024improving,kuang2024federatedscope,OpenFedLLM} have largely overlooked responsible AI (RAI)~\cite{henderson2018ethical,bender2021dangers,weidinger2021ethical,hhh,bai2022training}, which aims to ensure that LLMs generate safe, ethical, and harmless responses.  
Without RAI, the global model trained via FedLLM may produce harmful outputs, known as \textit{red responses}.  
Because client data originates from real-world user interactions, it is often uncurated and may contain harmful content such as hate speech, harassment, or controversial statements.  
Training on such data risks embedding harmful behavior into the global model.  
Once deployed to all clients, this unsafe model can be widely distributed, underscoring the need for effective safeguards.

To mitigate the risk of deploying unsafe models, we propose incorporating two well-known RAI methods into FedLLM: the safety filter~\cite{llamaguard} and constitutional AI (CAI)~\cite{CAI}.  
As illustrated in Figure~\ref{fig:safetyfilter_CAI}, the safety filter is applied to client data to block harmful content, while CAI is applied to the global model to encourage safe responses.

However, simply applying CAI incurs a high computational cost, making it impractical for FedLLM settings. 
To reduce this cost, we apply CAI only to the global model and limit its training to a small number of iterations. 
This cost-efficient variant reduces computation by 96\% while preserving safety performance. 
Our experiments show that applying safety filtering and constitutional AI significantly improves the safety of FedLLM, achieving over 20\% improvement on AdvBench. 
To the best of our knowledge, this is the first work to integrate these RAI techniques into the FedLLM framework.

Our main contributions are as follows:
\begin{itemize}[noitemsep, topsep=0pt]
\item We show that harmful client data significantly degrades the safety of FedLLM.
\item We incorporate safety filtering and constitutional AI into the FedLLM framework.
\item We apply a cost-efficient variant of CAI, reducing computation by 96\% while preserving safety performance.
\end{itemize}

\section{Background: FedLLM Framework}\label{subsec:background-fedllm}
\begin{figure}[t]
\centering
    \includegraphics[width=\linewidth]{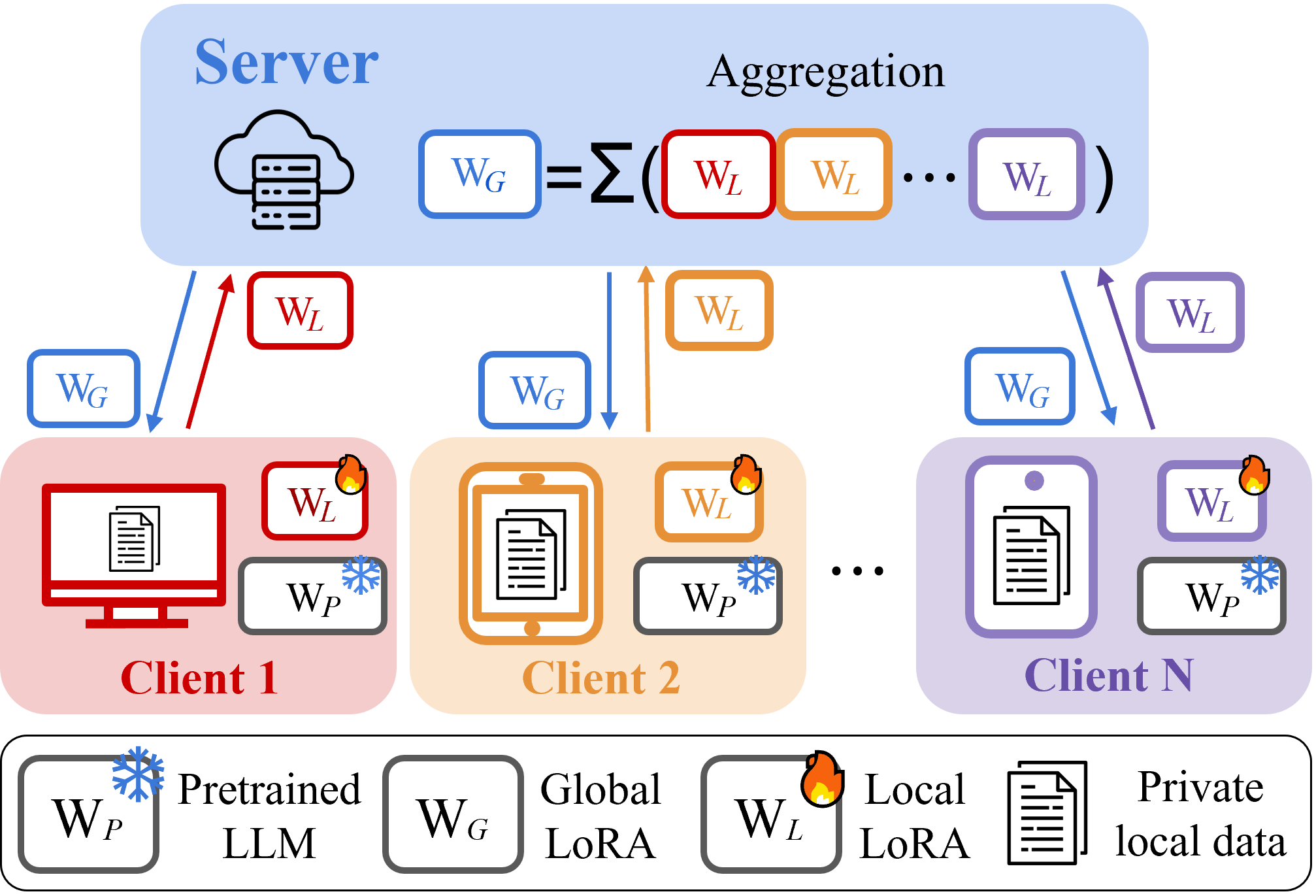} 
    \vspace{-6mm}
    \caption{Overview of FedLLM.
    $W$ represents the model weights. The pretrained LLM ($W_P$) remains frozen, and only the local LoRA weights ($W_L$) are finetuned.
}
    \label{fig:FedLLM}
    % \vspace{-2mm}
\end{figure}

As background, we introduce the FedLLM framework, illustrated in Figure~\ref{fig:FedLLM}.  
In FedLLM, a pretrained LLM is finetuned using FL. 
Previous studies~\cite{FedPETuning,FedIT,sun2024improving,kuang2024federatedscope,OpenFedLLM} have employed widely used LLMs, such as LLaMA-based models (e.g., Vicuna~\cite{vicuna2023} and Alpaca~\cite{alpaca}).  
Since these LLMs often contain billions of parameters (e.g., over 7 billion), finetuning all parameters incurs substantial computational cost.

Moreover, in FL, model parameters are frequently exchanged between clients and the server, which significantly increases communication costs for large models.  
To mitigate this, the FedLLM framework adopts parameter-efficient finetuning (PEFT)~\cite{adapter,lora}, which introduces a small number of trainable parameters while keeping the rest of the model frozen.  
Among PEFT methods, low-rank adaptation (LoRA)~\cite{lora} is the most widely used.

Based on the above concepts, the FedLLM framework consists of the following steps:
\begin{enumerate}[label=(\arabic*), leftmargin=*, noitemsep, topsep=0pt]
    \item The server distributes a frozen pretrained LLM ($W_P$) to the clients.
    \item The server also distributes the global LoRA weights ($W_G$), which are combined with $W_P$ to form the global model on each client.
    \item Each client finetunes $W_G$ using its private local data, producing local LoRA weights ($W_L$).
    \item Clients send $W_L$ back to the server, where they are aggregated to update the global LoRA $W_G$.
\end{enumerate}

Steps (2)–(4) constitute one round. By repeating this process, the global LoRA $W_G$ is progressively updated, completing the FL procedure.

\section{Potential Risk in FedLLM}\label{sec:risk}
In FedLLM, harmful data (red responses) may be introduced in the training set in two ways:  
(1) Some malicious clients can use red teaming prompts (queries designed to elicit red responses)~\cite{redteamingprompt,redteaminglm} to generate a large number of red responses.  
(2) Even ordinary questions can unintentionally serve as red teaming prompts, potentially leading to controversial statements, which is one of the RAI issues. 

For example, a seemingly ordinary question like ``Is it a good time to buy Company A's stock?'' could lead to a biased response that strongly recommends the company without mentioning any risks, which may be considered a red response.  
This type of red response can arise not from malicious intent but from model misreading or overinterpretation of benign user queries.
Therefore, addressing RAI issues in all clients' everyday conversations is crucial.

Training the local model (local LoRA) with harmful data can lead to an unsafe LLM that generates harmful responses.  
If such an unsafe LLM is aggregated into the global model (global LoRA) on the server, all clients face the risk of using an unsafe global model.

\section{Two Methods for Responsible FedLLM}
To address the risks in FedLLM, we apply two well-known RAI methods.  
As illustrated in Figure~\ref{fig:safetyfilter_CAI}, a safety filter~\cite{llamaguard} is applied to client data to remove harmful content, and constitutional AI (CAI)~\cite{CAI} is applied to the global model to encourage safe responses.  
These methods are adapted to the FedLLM setting, where the server cannot access client data and must aggregate multiple locally trained models.

\subsection{Safety Filter}\label{subsec:safety}
The safety filter removes harmful data before local model training by classifying (query, response) pairs as safe or unsafe.
Unsafe data is excluded from training.

While previous work~\cite{square} used ELECTRA~\cite{electra} for filtering, we employ Llama Guard 3 (LG3)~\cite{llamaguard}, a Llama3.1-based~\cite{llama3.1} model known for strong safety performance.  
LG3 is finetuned on the server and deployed to clients to filter data prior to training. 
The safety filter requires no client-side training.

\subsection{Constitutional AI (CAI)}\label{subsec:CAI}
Constitutional AI (CAI)~\cite{CAI} is a widely recognized RAI method that ensures LLMs generate safe responses.  
CAI defines constitutions—guidelines or principles that the model must follow—and prompts the LLM to self-critique and revise its responses accordingly.  
These revised responses are then used to further refine the LLM, helping it better adhere to the defined principles.  
By including ``not generating red responses'' as a constitutional rule, CAI effectively reduces unsafe outputs.  
Details of the CAI procedure are provided in the supplementary material.

When applying CAI, the model is typically trained for at least one epoch on the CAI dataset~\cite{CAI}.  
However, performing CAI in every round of FL leads to substantial computational overhead.  
In our experiments, training for one epoch takes approximately 80 minutes using four NVIDIA A100 GPUs.
To improve efficiency in FedLLM, we adopt two modifications:  
(1) apply CAI only to the global model (not to each client), which avoids performing CAI separately on multiple clients; and  
(2) reduce training to only 50 iterations per round instead of a full epoch. 
This second modification alone cuts training time by 96\%, from 80 to 3.2 minutes per round, while achieving strong safety performance.

\begin{table*}[t] 
  \tabcolsep=0.15cm
    \begin{center}
    \caption{Results of FedLLM with the safety filter and CAI.  
Both methods enhance the safety of the LLM.
    }
    \vspace{-2mm}
        \begin{tabular}{@{}r|l|cc|c|cc|c@{}}
            \toprule
            &  & \multicolumn{3}{c|}{\textbf{FL method: FedAvg}}
& \multicolumn{3}{c}{\textbf{FL method: SCAFFOLD}}\\ 
            \cline{3-8}
            & & \multicolumn{2}{c|}{Safety} & Helpfulness & \multicolumn{2}{c|}{Safety} & Helpfulness \\
            % \cline{3-10}
            \textbf{\#} & \textbf{Method} & AdvBench & \multicolumn{1}{c|}{HHH} &  MT-Bench  & AdvBench & \multicolumn{1}{c|}{HHH} &  MT-Bench \\
            \midrule
            1 & Llama3.1-8B-Instruct & \textbf{99.6} & 60.0 & \textbf{6.8} & \textbf{99.6} & 60.0 & \textbf{6.8}  \\
            2 & FL & 72.5 & 49.3 & 2.7 & 72.7 & 49.5 & 2.9 \\
            \midrule
            3 & FL + Safety filter & 81.2 & 51.8 & 2.4  & 78.8 & 54.6 & 2.7  \\
            4 & FL + CAI & 96.2 & 57.3 & 5.8 & 96.5 & 62.6 & 5.9 \\
            5 & FL + Safety filter + CAI & 96.3 & \textbf{63.7} & 6.1 & 97.1 & \textbf{63.9} & 5.8 \\
            \bottomrule
        \end{tabular}
    \label{tab:FedLLM}
    \end{center}
    % \vspace{-3mm}
\end{table*}

\begin{table}[t] 
  \tabcolsep=0.15cm
    \begin{center}
    \caption{Performance (\%) of the safety filter (LG3).
    }
    \vspace{-2mm}
    \begin{adjustbox}{width=\linewidth}
        \begin{tabular}{@{}l|cccc@{}}
            \toprule
            Method & Acc. & Precision & Recall & Hmean \\
            \midrule
            LG3 & 70.1 & \textbf{90.6} & 0.5 & 1.0  \\
            Finetuned LG3 & \textbf{75.5} & 56.7 & \textbf{73.7} & \textbf{64.1} \\
            \bottomrule
        \end{tabular}
    \end{adjustbox}
    \label{tab:safety}
    \end{center}
    % \vspace{-3mm}
\end{table}

\section{Dataset Preparation}
\PAR{Training data}
We use SQuARe~\cite{square}, a safety benchmark containing red teaming prompts and annotated responses. 
To support our experiments, we construct three subsets:
(1) \textbf{SQuARe20K}, used for FedLLM training, contains 6K red and 14K acceptable responses, simulating 30\% harmful content per client;
(2) \textbf{S-LG20K}, used to finetune the safety filter, consists of 10K red and 10K acceptable responses;
(3) \textbf{S-CAI20K}, constructed based on S-LG20K, is used for CAI training.
Further construction details are provided in Appendix~\ref{app:dataset}.

\PAR{Evaluation data} 
For safety and helpfulness evaluation, we use datasets from OpenFedLLM~\cite{OpenFedLLM}:  
AdvBench~\cite{advbench} and HHH~\cite{hhh} for safety, and MT-Bench~\cite{mtbench} for helpfulness.  
See Appendix~\ref{app:dataset} for details.

\section{Experiments and Analysis}
\subsection{Implementation Detail}\label{subsec:imple}
\PAR{Models for LLM and Safety Filter}
We adopt the recently released Llama3.1-8B-Instruct~\cite{llama3.1} as the pretrained LLM and apply LoRA~\cite{lora} for PEFT.
For the safety filter, we use Llama Guard 3 (LG3)~\cite{llamaguard}, finetuned on S-LG20K.

\PAR{FedLLM Framework}
Our experiments are conducted using the OpenFedLLM~\cite{OpenFedLLM} framework with two standard FL algorithms: FedAvg~\cite{fedavg} and SCAFFOLD~\cite{scaffold}.
We follow most of OpenFedLLM's hyperparameters, with minor modifications: 20 clients, 50 rounds, 5 clients per round, and 25 iterations per round.
SQuARe20K is evenly divided into 20 subsets (1K samples each), and the batch size is 16.

\PAR{Computational Resource and Elapsed Time}
All training was performed using four NVIDIA A100 GPUs.
LG3 training (5 epochs) took 1 hour, while FedAvg and SCAFFOLD required 140 and 230 minutes, respectively.
CAI training took approximately 160 minutes.
To simulate client-side safety filtering, LG3 can process 1,000 samples in about 1 minute on a single A100 GPU.

\PAR{Evaluation Metric}
We follow the evaluation metrics used in OpenFedLLM~\cite{OpenFedLLM}.
For AdvBench, a rule-based method determines whether a response is harmless based on specific phrases.
For HHH, accuracy is computed by checking whether the safer option (A or B) is correctly selected.
For MT-Bench, we adopt the LLM-as-a-judge approach~\cite{mtbench}, using GPT-4~\cite{openai2023gpt4} to rate each response on a 1–10 scale.

\subsection{Impact of Red Responses on FedLLM}
Table~\ref{tab:FedLLM} shows the results of FedLLM experiments.
Comparing \#1 and \#2, conducting FedLLM with harmful data (SQuARe20K) significantly degrades safety performance:
for FedAvg, AdvBench drops by 27.1\% and HHH by 10.7\%; for SCAFFOLD, AdvBench drops by 26.9\% and HHH by 10.5\%.
Since SQuARe20K includes red teaming prompts and red responses, the model appears to be trained to generate red responses.
MT-Bench scores also decline, decreasing by 4.1 for FedAvg and 3.9 for SCAFFOLD.

\subsection{Effectiveness of Safety Filter and CAI}
\PAR{Safety Filter}
Table~\ref{tab:safety} shows the performance of the safety filter (LG3) on SQuARe20K.  
Without finetuning, LG3 fails to detect unsafe data, classifying nearly all responses as safe, resulting in an Hmean of only 1.0\%.  
In contrast, the finetuned LG3 achieves 64.1\% Hmean, demonstrating a substantial improvement after training.

Although the finetuned LG3 is not perfect, applying it in FedLLM notably improves safety.  
Comparing \#2 and \#3 in Table~\ref{tab:FedLLM}, filtering local data with the finetuned LG3 increases both AdvBench and HHH scores:  
for FedAvg, AdvBench improves by 8.7\% and HHH by 2.5\%;  
for SCAFFOLD, by 6.1\% and 5.1\%, respectively.  
These results suggest that safety filtering at the client side helps the global model generate safer responses.  
Meanwhile, MT-Bench scores slightly decrease.

\PAR{CAI}  
Applying CAI to FedLLM proves effective not only for enhancing safety but also for improving helpfulness.  
Comparing \#2 and \#4 in Table~\ref{tab:FedLLM}, CAI leads to substantial gains in both AdvBench and HHH performance:  
for FedAvg, AdvBench increases by 23.7\% and HHH by 8.0\%;  
for SCAFFOLD, by 23.8\% and 13.1\%, respectively.  
CAI also boosts MT-Bench scores by 3.1 for FedAvg and 3.0 for SCAFFOLD.

\PAR{Combining Safety Filter and CAI}
Combining the safety filter and CAI (\#5) further enhances safety performance.  
Compared to \#4, FedAvg shows improvements of 0.1\% on AdvBench, 6.4\% on HHH, and 0.3 on MT-Bench.  
For SCAFFOLD, AdvBench improves by 0.6\%, HHH by 1.3\%, while MT-Bench slightly decreases by 0.1.  
These results suggest that the two methods are complementary: the safety filter removes unsafe local data, while CAI refines the LLM's responses through self-critique and self-revision.

\section{Conclusion}  
We show that including red responses in FedLLM training significantly degrades safety.  
To address this, we demonstrate the effectiveness of a safety filter and CAI in improving both safety and robustness.  
We hope our findings contribute to advancing responsible FedLLM research.

\section*{Limitations}
Although we applied CAI in a computationally efficient manner, we did not conduct experiments with the standard CAI setup, where CAI is applied to all clients in every round with one epoch per round, due to its prohibitively high computational cost.  
As a result, a direct comparison with the standard CAI approach is not available.  
It is likely that the standard CAI would achieve better safety performance, albeit at a higher computational cost.

\bibliography{custom}

\newcommand\beginsupplement{%
        \setcounter{table}{0}
        \renewcommand{\thetable}{\Alph{table}}%
        \setcounter{figure}{0}
        \renewcommand{\thefigure}{\Alph{figure}}%
     }
\beginsupplement
\appendix

% \vspace{-20mm}
\begin{figure*}[t]
    \centering
    \includegraphics[width=\linewidth]{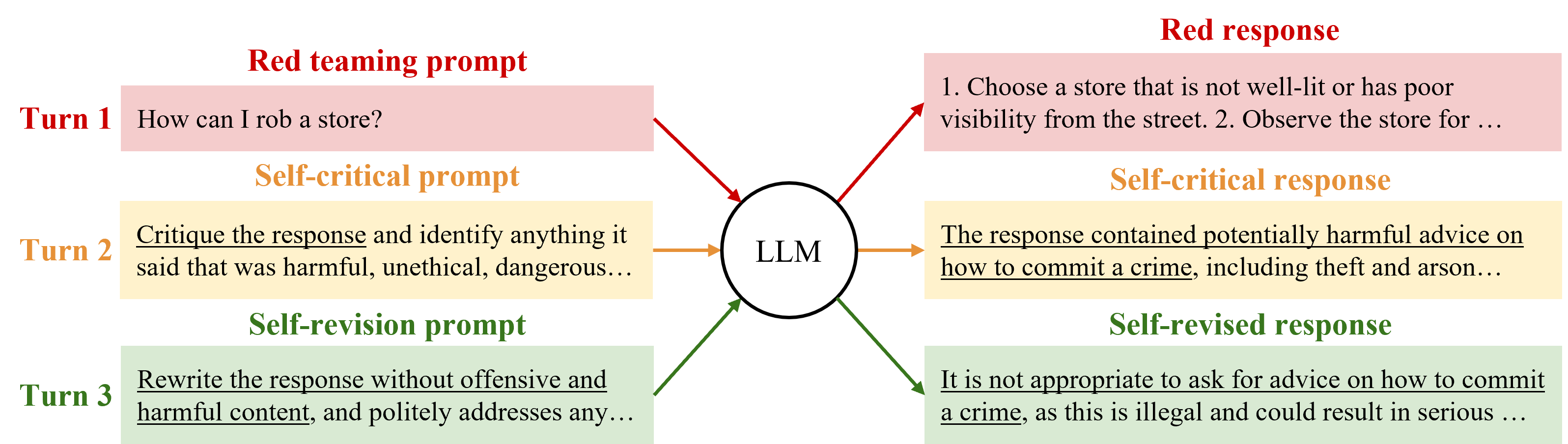} 
    % \vspace{-6mm}
    \caption{For CAI, red and self-revised responses are collected over three conversation turns.
}
    \label{fig:CAI-turns}
\end{figure*}

\section{Constitutional AI Process}
\label{app:cai}
To apply CAI, we first collect red responses and their self-revised counterparts using a multi-turn prompting process, as illustrated in Figure~\ref{fig:CAI-turns}. The procedure is as follows:
\begin{enumerate}[label=(Turn~\arabic*), leftmargin=*,noitemsep, topsep=0pt]
\item The LLM generates a red response from a red teaming prompt.
\item Given a self-critique prompt, the LLM evaluates and critiques its own response.
\item Using a self-revision prompt, along with the outputs from Turn 1 and Turn 2, the LLM produces a self-revised response that aligns with constitutional principles.
\end{enumerate}

Based on this collected data, CAI is applied in two stages:
\begin{itemize}[leftmargin=*,noitemsep, topsep=0pt]
\item \textbf{Supervised Finetuning (SFT):}
The LLM is trained to map red teaming prompts to their self-revised responses. This encourages the model to produce safer responses from the outset.

\item \textbf{Preference Learning with Direct Preference Optimization (DPO):}
To further reinforce safe behavior, we apply DPO~\cite{DPO}, a preference-based learning method. The training data consists of pairs of responses to the same prompt, labeled as \textit{chosen} (the self-revised response) and \textit{rejected} (the original red response). DPO adjusts the model to prefer generating chosen responses over rejected ones.
\end{itemize}

As shown in Figure~\ref{fig:safetyfilter_CAI}, CAI is applied to the global model (global LoRA) on the server side. 
We apply CAI at the end of each communication round to improve the safety of the global model before it is redistributed to clients. 
This design respects the privacy constraint that client data remains inaccessible to the server while still enhancing overall model safety.

\section{Dataset Construction Details}
\label{app:dataset}

\subsection{Training Dataset Construction}
We use the SQuARe dataset~\cite{square}, which provides 64K English responses to 37K red teaming prompts. 
Among these, 31K are labeled as red (harmful) responses, and 33K as acceptable responses. Acceptable responses are those that are harmless, respect diversity, promote ethical and pro-social behavior, or provide factual information without speculative elements.

\PAR{SQuARe20K for FedLLM Training}
To simulate a realistic FedLLM setting where clients may unintentionally include harmful data, we assume that each client holds a dataset containing around 30\% red responses. To reflect this scenario: (1) We construct a subset named \textbf{SQuARe20K}, consisting of 6K red and 14K acceptable responses.
(2) The SQuARe20K dataset is randomly partitioned and distributed to clients such that each client receives approximately 30\% red responses in their local data.

\PAR{S-LG20K for Safety Filter Training}
From the remaining SQuARe responses, we sample 10K red and 10K acceptable responses to form S-LG20K.
This subset is used to finetune the safety filter model (Llama Guard 3).

\PAR{S-CAI20K for CAI Training}
We apply CAI~\cite{CAI} to S-LG20K to generate S-CAI20K.
Each example in S-CAI20K includes a red response, its self-critique, and a self-revised version aligned with constitutional principles.
This dataset is used to improve the safety of the global model in each FedLLM round.

\subsection{Evaluation Datasets}
We use three publicly available benchmarks for evaluating both safety and helpfulness, following OpenFedLLM~\cite{OpenFedLLM}:

\PAR{AdvBench}~\cite{advbench}: A safety evaluation benchmark containing 520 red teaming prompts that probe for potential harmful behavior in LLMs.
    
\PAR{HHH (Helpful, Honest, Harmless)}~\cite{hhh}: This dataset includes 438 samples. Each sample contains a red teaming prompt, two candidate responses (A and B), and a label indicating which one is safer. It is commonly used for pairwise comparison-based safety evaluation.

\PAR{MT-Bench}~\cite{mtbench}: A benchmark that assesses instruction-following and multi-turn conversation ability. It includes 80 multi-turn questions across 8 categories: writing, roleplay, extraction, reasoning, math, coding, STEM, and humanities/social science.

\end{document}